\documentclass{article}
\usepackage{spconf}
\usepackage{amsmath}
\usepackage{amsfonts}
\usepackage{amssymb}
\usepackage{algorithmic}
\usepackage{graphicx}
\usepackage{textcomp}
\usepackage{xcolor}
\usepackage{multirow,multicol}

\title{ ADAPATIVE STEP-SIZE PERCEPTION UNFOLDING NETWORK WITH NON-LOCAL HYBRID ATTENTION FOR HYPERSPECTRAL IMAGE RECONSTRUCTION}

\name{YaNan Yang,Like Xin*\thanks{Like Xin is with the School of Mathematical Sciences, Nanjing Normal University, Nanjing, 210046, China.  \protect (e-mail: xinlike94@gmail.com).}}

\address{Nanjing Normal University}

\begin{document}

\maketitle
\begin{abstract}
Deep unfolding methods and transformer architecture have recently shown promising results in hyperspectral image (HSI) reconstruction. However, there still exist two issues: (1) in the data subproblem, most methods represents the step-size utilizing a learnable parameter. Nevertheless, for different spectral channel, error between features and ground truth is unequal. (2) Transformer struggles to balance receptive field size with pixel-wise detail information. To overcome the aforementioned drawbacks, we proposed an adaptive step-size perception unfolding network (ASPUN), a deep unfolding network based on FISTA algorithm, which uses an adaptive step-size perception module to estimate the update step-size of each spectral channel. In addition, we design a Non-local Hybrid Attention Transformer(NHAT) module for fully leveraging the receptive field advantage of transformer. By plugging the NLHA into the Non-local Information Aggregation (NLIA) module, the unfolding network can achieve better reconstruction results. Experimental results show that our ASPUN is superior to the existing SOTA algorithms and achieves the best performance.
\end{abstract}
\begin{keywords}
Hyperspectral image reconstruction, FISTA algorithm, non-local attention, Deep unfolding network.
\end{keywords}
\section{Introduction}
\label{sec:intro}

hyperspectral image can obtain detailed spectral information of objects or scenes~\cite{ref1,ref2}, widely used in agricultural monitoring, geological exploration, medical diagnosis,etc. CASSI technique provides an efficient approach for acquiring spectral data. Nonetheless, the reconstruction of an accurate and detailed 3D hyperspectral image (HSI) cube from the 2D measurements poses a fundamental challenge for the CASSI system.\par
Based on CASSI, various reconstruction techniques have been developed to reconstruct the 3D HSI cube from 2D measurements. These methods range from traditional model-based methods to deep learning based methods. Model-based HSI reconstruction methods~\cite{ref3,ref4,ref5,ref6} commonly optimize the objective function by separating the data fidelity term and regularization term. Twist~\cite{ref3}  introduced a two-step Iterative shrinkage/thresholding algorithm for reconstruction with missing samples. The non-local similarity and low-rank regularization were utilized in ~\cite{ref4}  and ~\cite{ref5}. Although these methods produce satisfying results in the case of proper situation, they still face the problems of lacking generalization ability and computational efficiency.
Thanks to fast developments of attention mechanism~\cite{ref7}, Tansformer based methods~\cite{ref8,ref9} have gained widespread utilization in the Artificial intelligence.To explore the power of Transformer for compressive sensing,MST [10] and CST [11] were proposed to capture the inner similarity of HSIs. Hu et al. [12] introduced the high resolution dual-domain learning network (HDNet) to solve the spectral compressive imaging task. Though significant progress has been made, it is difficult for these brute-force methods to utilize the physical degradation characteristics.However, these methods have certain limitations in an open environment.
Yang et al.t ~\cite{ref20} propose PAL, a sampling scheme that progressively selects valuable OOD instances by evaluating their informativeness and representativeness, balancing pseudo-ID and pseudo-OOD instances to enhance the ID classifier and OOD detector. Yang et al.t~\cite{ref22} present WiseOpen, an OSSL framework that uses a gradient-variance-based selection mechanism to enhance ID classification by selectively leveraging a favorable subset of open-set data for training.
 \par
These methods achieve good performance to some extent, but still have many problems and can be summarized in two key aspects:(1)Typically, the deep unfolding methods address a data subproblem and a prior subproblem iteratively.In the data subproblem, most methods use a learnable parameter to represent the step size of the iteration in gradient descent, however, each channel contributes differently to the loss, and the optimal iteration step size should be different for different channels. Moreover, this problem is particularly obvious in hyperspectral reconstruction problems with a large number of channels. (2)In the case of the same computational cost, it is difficult for the method based on transforemer to obtain both large receptive field and accurate pixel information. If the size of the receptive field is increased, the accuracy of the information will be sacrificed, which restricts the reconstruction effect of the model. How to effectively utilize the advantages of transformer remote similarity modeling has become a challenging problem.\par

\begin{figure*}[!ht]
    \centering
    \setlength{\abovecaptionskip}{0.cm}
    \includegraphics[width=1\textwidth]{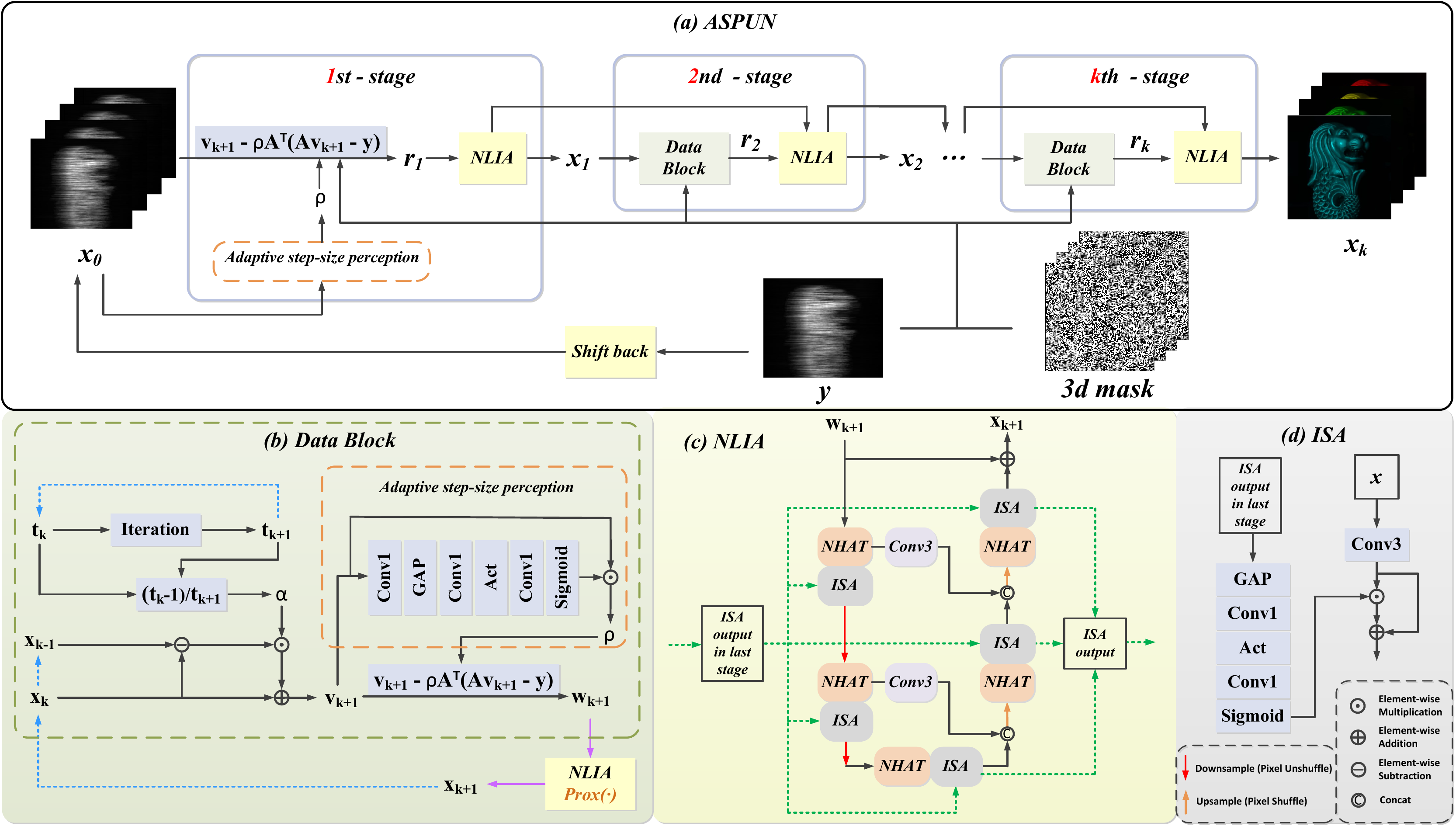}
    \caption{{Overall architecture of our designed adaptive step-size perception unfolding network (ASPUN)}}
    \end{figure*}
    
To solve the above issues, We propose an adapative step-size perception unfolding  network(ASPUN) based on FISTA algorithm, which uses an adaptive step perception module to estimate the update step of each channe to obtain a better reconstruction effect.Morever,a novel module for hyperspectral image reconstruction is proposed, we call our moudle as Non-local Information Aggregation (NLIA), which adopts a U-shaped structure consists of several basic unit NHAT blocks. As a major component of NHAT ,NLHA is a vison transformer that can effectively capture local and non-local information, At the same time, a gated local attention module is used to make up for the lack of information accuracy. Equipped with the proposed techniques,ASPUN-NLHA achieves state-of-the-art (SOTA) performance on HSI reconstruction.

\section{Method}
\label{method}
\subsection{Problem Formulations}
Using a fixed mask and a disperser to achieve band modulation, the Coded Aperture Snapshot Spectral Imaging (CASSI) system can compress 3D data cubes$(x,y,\lambda)$into snapshot measurements on a 2D detector.
The physical mask,denoted by ${M\in\mathbb{R}^{H \times W}}$, acts as a modulator for the HSI signal $X \in \mathbb{R}^{H\times W\times N_\lambda}$, thereby enabling the representation of the $nth_{\lambda}$ wavelength of the modulated image:
\begin{align}
X^{\prime}{\big(}:,:,n_{\lambda}{\big)} = X{\big(}:,:,n_{\lambda}{\big)}\odot M
\end{align}
where $\odot$represents element-wise multiplication.Consequently, the modulated HSI X are shifted during the dispersion process, which can be expressed as:
\begin{align}
X^{\prime \prime}{\big(}u,v,n_{\lambda}{\big)} = X^{\prime}{\big(}x,y+d{\big(}{\lambda}_n- {\lambda}_c{\big)},n_{\lambda}{\big)}
\end{align}
where (u, v) indicates the coordinate system on the detector plane, and $\lambda_n$ is the wavelength of channel $n_\lambda$. Here, $d{\big(}{\lambda}_n- {\lambda}_c{\big)}$ signifies the spatial shifting for channel $n_\lambda$. 
At last, the imaging sensor captures the shifted image into a 2D measurement. This process can be formulated as follow:
\begin{align}
Y = \sum_{n_\lambda = 1}^{N_\lambda} X^{\prime \prime }(:,:,n_\lambda)
\end{align}
the 3D HSI cube is degraded to 2D measurement $Y \in \mathbb{R}^{H\times (W+d_N\lambda)}$after the sum operator, and the spatial dimensions increased as the dispersion process. 

\subsection{Adaptive Step-size Perception Unfolding Network}

Considering the measurement noise, the matrix-vector form of Eq(3). can be formulated as:
\begin{align}
y = \Phi x+n
\end{align}
\begin{figure}[!ht]
    \centering
    \setlength{\abovecaptionskip}{0.cm}
    \includegraphics[width=0.5\textwidth]{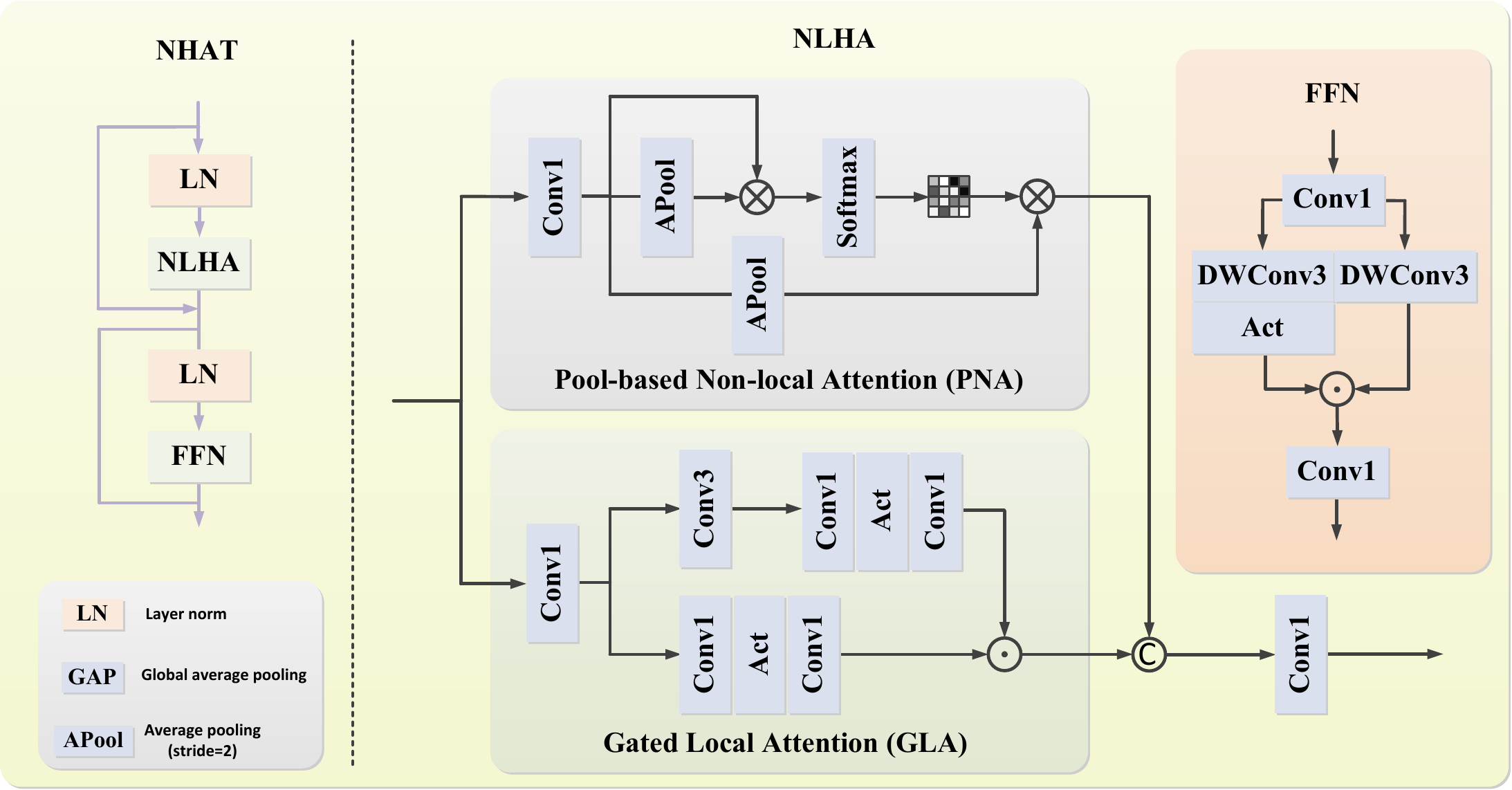}
    \caption{{Illustration of our designed Non-local Hybrid Attention Transformer.}}
    \label{NHAT}
    \end{figure}
where x is the original HSI, y is the degraded measurement,$\Phi$ is the sensing matri x, and n represents the additive noise. Conventional methods usually employ a regularization term R(x) as prior to constrain the solution in desired signal space. These algorithms aim to find an estimated $\bar{x}$ of x by solving the following problem:
\begin{align}
\bar{x}=\arg\min_x\frac12||y-\Phi x||_2^2+\lambda R(x)
\end{align}
where $\lambda$ is a parameter to balance between the fidelity and the regularization term. Eq(5). is usually solved by iterative algorithms with various image priors of R(x)including sparsity, total variation, deep denoising prior, autoencoder prior, etc.
Applying the sparsity constraint as the regular term in Eq(5), The reconstruction problem can be abstracted as following optimization problem:
\begin{align}
\min_{{x}}\frac{1}{2}\|{\Phi}x-y\|_2^2+\lambda\|{\Psi}x\|_1
\end{align}
where ${\Psi}x$ denotes the coefficients in transformation domain, and $\lambda$ represents a balanced coefficients.
By optimizing $x_k$ and $r_k$ alternately, FISTA algorithm attempt to solve the optimization problem by performing following steps iteratively:
\begin{gather}
r^{k} ={z}^k-\rho{\Phi}^T({\Phi}{z}^k-{y}), \\
x^{k} =\arg\min_{{x}}\frac12\|{x}-{r}^k\|_2^2+\lambda\|{\Psi}{x}\|_1 \\
t^{k+1} =\frac{1+\sqrt{1+4(t^k)^2}}2, \\
{z}^{k+1} =x^k+(\frac{t^k-1}{t^{k+1}})(x^k-x^{k-1}), 
\end{gather}

where, $z_k$ is an auxiliary variable representing the starting point of k-th iteration, and $\rho$ represents the step size, usually implemented by a learnable parameter. 
Traditional FISTA algorithm using only one learnable parameter to characterize the step size hyperparameter results in each spectral channel equally receiving the error between the image degraded by $z_k$ and the real measurement, from which the $r_k$ is updated by gradient descent. 
\subsection{Non-local Information Aggregation}
 As shown in Fig~\ref{NHAT}NLIA module adopt a U-shaped architecture trying to solve optimization step in eq .   It consists of five NHATs and ISAs.   NHAT is responsible for capturing and reconstructing spatial-wise pixel details, while ISA learns a set of spectral-wise coefficients from previous stage to suppress unimportant spectral channel information.\\
\textbf{Non-local Hybrid Attention Transformer.} NHAT is designed following the metaformer architecture.   We embed two layer normalization, a NLHA and a gated FFN in NHAT with two skip connections. NLHA reconstructs image features from both non-local and local perspectives, while FFN act as a multilayer linear perceptron.\\
\textbf{NLHA.} In NLHA, we use a PNA module to model non-local self attention. It introduces pooling operation in the original transformer architecture, which enables PNA to have a larger receptive field and more effectively perceive non-local structural information of images at the same computational cost.  However, within each patch of the PNA, information is exchanged between all tokens, and excessive non-local information propagation reinforces the low-frequency representation, which may destroy the detailed texture of the image.  In addition, the pooling operation in PNA exacerbates the destruction of image details, which will degrade the performance of vision transformer.  Considering the insufficient ability of transformer to process local information, we designed a cnn-based module GLA to .  It uses the local perception and weight sharing characteristics of convolution kernel to sense local context information and generate fine image details and textures, which is complementary to PNA.  Specifically, GLA adopts a two-branch structure, where one branch uses a 3x3 vanilla convolution as a local context sensor, and then after a simple nonlinear point convolution operation on the features of both branches, the context branch uses a sigmoid function to generate a local attention score.  The two branches then pass through the hadamard product to generate local information.

\section{Experiment}
\subsection{Experimental Settings}
Following the previous work, we used 28 wavelengths in the range of 450nm to 650nm to simulate and perform real experiments on HSI. These wavelengths are derived by a spectral interpolation operation.\\
\textbf{Datasets.} For simulation experiments, We train on the CAVE dataset and test on 10 scenes selected from the KAIST dataset. Besides, we also test the performance of our algorithm on real measurements captured by real CASSI system.\\
\begin{table*}[!ht]
\begin{center}
\setlength{\abovecaptionskip}{0.cm}
\caption{{Comparison of our methods and the other approaches in 10 simulated scenes (s1-s10). PSNR/SSIM metrics are reported.}} \label{table:performance}
\resizebox{\textwidth}{!}{
\begin{tabular}{c|ccccccccccccc}
\hline
  Framework & Params & flops(G) & s1 & s2 & s3 & s4 & s5 & s6 & s7 & s8 & s9 & s10 & avg
  \\  \hline
  \multirow{2}*{\centering TwIsT} & \multirow{2}*{\centering -} & \multirow{2}*{\centering -} & 25.16 & 23.02 & 21.40 & 30.19 & 21.41 & 20.95 & 22.20 & 21.82 & 22.42 &  22.67 & 23.12\\ 
   & &  & 0.700 & 0.604 & 0.711 & 0.851 & 0.635 & 0.644 & 0.643 & 0.650 & 0.690 &  0.569 & 0.669\\  \hline
  \multirow{2}*{\centering GAP-TV} & \multirow{2}*{\centering -} & \multirow{2}*{\centering -} & 26.82 & 22.89 & 26.31 & 30.65 & 23.64 & 21.85 & 23.76 & 21.98 & 22.63 & 23.10 & 24.36 \\
   &  &  &  0.754 & 0.610 & 0.802 & 0.852 & 0.703 & 0.663 & 0.688 & 0.655 & 0.682 & 0.584 & 0.669 \\  \hline
 \multirow{2}*{\centering DeSCI} & \multirow{2}*{\centering -} & \multirow{2}*{\centering -} & 27.13 & 23.04 & 26.62 & 34.96 & 23.94 & 22.38 & 24.45 & 22.03 & 24.56 & 23.59 & 25.27 \\
   & &  & 0.748 & 0.620 & 0.818 & 0.897 & 0.706 & /0.683 & 0.743 & 0.673 & 0.732 & 0.587 & 0.721 \\  \hline
  \multirow{2}*{\centering $\lambda$-Net} & \multirow{2}*{\centering 62.64M} & \multirow{2}*{\centering 117.9} & 30.10 & 28.49 & 27.73 & 37.01 & 26.19 & 28.64 & 26.47 & 26.09 & 27.50 & 27.13 & 28.53 \\
   & &  & 0.849 & 0.805 & 0.870 & 0.934 & 0.817 & 0.853 & 0.806 & 0.831 & 0.826 & 0.816 & 0.841 \\  \hline
  \multirow{2}*{\centering TSA-Net} & \multirow{2}*{\centering 44.25M} & \multirow{2}*{\centering 110.0} & 32.03 & 31.00 & 32.25 & 39.19 & 29.39 & 31.44 & 30.32 & 29.35 & 30.01 & 29.59 & 31.46 \\ 
   & &  & 0.892 & 0.858 & 0.915 & 0.953 & 0.884 & 0.908 & 0.878 & 0.888 & 0.890 & 0.874 & 0.894 \\  \hline
  \multirow{2}*{\centering HDNet} & \multirow{2}*{\centering 2.37M} & \multirow{2}*{\centering 154.7} & 35.14 & 35.67 & 36.03 & 42.30 & 32.69 & 34.46 & 33.67& 32.48 & 34.89 & 32.38 & 34.97 \\
   & &  & 0.935 & 0.940 & 0.943 & 0.969 & 0.946 & 0.952 & 0.926 & 0.941 & 0.942 & 0.937 & 0.943 \\  \hline
  \multirow{2}*{\centering BIRNAT} & \multirow{2}*{\centering 4.40M} & \multirow{2}*{\centering 2122.6} & 36.79 & 37.89 & 40.61 & 46.94 & 35.42 & 35.30 & 36.58 & 33.96 & 39.47 & 32.80 & 37.58 \\
   &  &  & 0.951 & 0.957 & 0.971 & 0.985 & 0.964 & 0.959 & 0.955 & 0.956 & 0.970 & 0.938 & 0.960 \\  \hline
  \multirow{2}*{\centering DAUHST-L} & \multirow{2}*{\centering 6.15M}  & \multirow{2}*{\centering 79.50} & 37.25 & 39.02 & 41.05 & 46.15 & 35.80 & 37.08 & 37.57 & 35.10 & 40.02 & 34.59 & 38.36 \\
   &  &  & 0.958 & 0.967 & 0.971 & 0.983 & 0.969 & 0.970 & 0.963 & 0.966 & 0.970 & 0.956 & 0.967 \\  \hline
  \multirow{2}*{\centering DADF-Plus-3} & \multirow{2}*{\centering 58.13M} & \multirow{2}*{\centering 230.4} & 37.46 & 39,86 & 41.03 & 45.98 & 35.53 & 37.02 & 46.76 & 34.78 & 40.07 & 34.39 & 38.29 \\
  &  &  & 0.965 & 0.976 & 0.974 & 0.989 & 0.972 & 0.975 & 0.958 & 0.971 & 0.976 & 0.962 & 0.972   \\  \hline
  \multirow{2}*{\centering RDLUF-9stg} & \multirow{2}*{\centering 1.89M} & \multirow{2}*{\centering -} & 37.94 & 40.95 & 43.25 & 47.83 & 37.11 & 37.47 & 38.58 & 35.50 & 41.83 & 35.23 & 39.57 \\
  &  &  & 0.966 & 0.977 & 0.979 & 0.990 & 0.976 & 0.975 & 0.969 & 0.970 & 0.978 & 0.962 & 0.974 \\ \hline
  \multirow{2}*{\centering ASPUN-9stg(ours)} & \multirow{2}*{\centering 10.77M} & \multirow{2}*{\centering 194.8} & 38.32 & 41.29 & 44.27 & 48.83 & 38.67 & 37.88 & 40.32 & 36.60 & 42.38 & 35.48 & 40.40 \\
  &  &  & 0.973 & 0.983 & 0.984 & 0.994 & 0.984 & 0.980 & 0.979 & 0.980 & 0.984 & 0.970 & 0.981 \\  \hline
\end{tabular}
}
\end{center}
\end{table*}
\textbf{Implementation Details.} We use the Adam optimizer and adjust the learning rate with Cosine Annealing schedule. The initial learning rate is $3x10^{-4}$ and Charbonnier loss isused for the loss function. our ASPUN-S, ASPUN-M and ASPUN-L are versions of ASPUN with 3, 6and 9 cascading NHIAs.\\ 
\textbf{Competing Methods.} We compare our method with 13state-of-the-art (SOTA) methods, which contain 3 traditional model-based methods(TwIST~\cite{ref3}, GAP-TV~\cite{ref15},DeSCI~\cite{ref16}) and 10 deep learning based methods($\lambda$-Net~\cite{ref17},TSA-Net~\cite{ref9}, BIRNAT~\cite{ref19}, DAUHST~\cite{ref14}, DADF~\cite{ref21} and RDLUF~\cite{ref18}).
\subsection{Comparisons with SOTA}
\textbf{Quantitative Comparisons.} Comparison results on ten simulation scenes are presented in Table~\ref{table:performance}. It is shown that our method outperforms all SOTA methods. Compared to HDNet, BIRNAT, DAUHST-L, and DADF-Plus-3, our ASPUN-9stg achieve improvements of 5.43dB, 2.62dB, 2.04dB, and 1.11dB respectively. In particular, ASPUN-9stg beats RDLUF-L with 0.83dB PSNR and 0.009 SSIM boosts.
\textbf{Qualitative Comparisons.} Fig.~\ref{simu} shows the visualization of simulated HSI reconstruction results on scene 7 using different methods. Our approach shows the best performance in spatial detail reconstruction. Especially in the area of the yellow box after magnification, our method correctly captures the underlying structural information of the image, restoring the most accurate and visually pleasing texture. This is due to the fact that our remote branches accurately capture the structure information of the image, while the local branches complement the detailed texture information of the image.

\begin{table}[!bt]
\begin{center}
\setlength{\abovecaptionskip}{0.cm}
\caption{{Ablation study on branches and components in NLHA.}} \label{table:ablation}
\resizebox{0.5\textwidth}{!}{
\begin{tabular}{c|c|c|c}
\hline
  branch & Methods & PSNR(dB) & SSIM 
  \\  \hline
  \multirow{2}*{\centering GLA} &  w/o context-aware & 38.56 & 0.973\\ 
     & w/o GLA branch & 38.08 & 0.971\\  \hline
  \multirow{2}*{\centering PNA} & w/o pool-based transformer &38.34 & 0.973\\
     &  w/o PNA branch& 38.14 &0.970\\  \hline
  \multicolumn{2}{c|}{complete ASPUN}& 38.85 & 0.976\\ \hline
\end{tabular}
}
\end{center}
\end{table}

\subsection{Ablation Study}
\textbf{Break-down Study.} To analyze the impact of each module on performance, we conducted a breakdown ablation study, the results of which are presented in the table. We removed the NLHA, ISA, and ASP modules from the original model to form baseline, which uses a single learnable parameter to measure step size, achieving 37.17dB. With the introduction of ASP, a more flexible step-aware strategy resulted in a 0.03dB performance improvement. Adding ISA and NLHA to reconstruct features from spectral and spatial dimensions increases performance by 0.50dB and 1.15dB, respectively. This proves the effectiveness of each module we design.
\begin{table}[!bt]
\begin{center}
\setlength{\abovecaptionskip}{0.cm}
\caption{{Performance comparison of different attention architectures.}} \label{table:ablation2}
\resizebox{0.5\textwidth}{!}{
\begin{tabular}{c|c|c|c|c|c}
\hline
  branch & Attention & Params(M) & FLOPs(G) & PSNR(dB) & SSIM
  \\  \hline
  \multirow{2}*{\centering w/o GLA} &  WMSA & 2.48 & 45.7 & 38.16 & 0.971 \\ 
     & PNA & 2.48 & 45.7 & 38.08 & 0.971 \\  \hline
  \multirow{2}*{\centering with GLA} & WMSA & 3.58 & 62.6 & 38.64 & 0.975 \\
     &  PNA & 3.58 & 62.6 & 38.85 & 0.976 \\  \hline

\end{tabular}
}
\end{center}
\end{table}
\begin{table}[!ht]
\begin{center}
\setlength{\abovecaptionskip}{-0.3cm}
\caption{Break-down ablation on ASP, ISA and NLHA.} \label{table:breakdown}
\resizebox{0.25\textwidth}{!}{
\begin{tabular}{c|cccc}
  \hline
  \quad & PSNR  & SSIM   \\ \hline
        baseline & 37.17 &  0.964   \\ \hline
        + \textit{ASP}& 37.20 & 0.964  \\ \hline
        + \textit{ISA}& 37.70 &  0.968  \\ \hline
        + \textit{NLHA}& 38.85 &  0.976 \\  \hline
\end{tabular}
}
\end{center}
\end{table}

\begin{figure*}[!ht]
    \centering
    \setlength{\abovecaptionskip}{0.cm}
    \includegraphics[width=\textwidth]{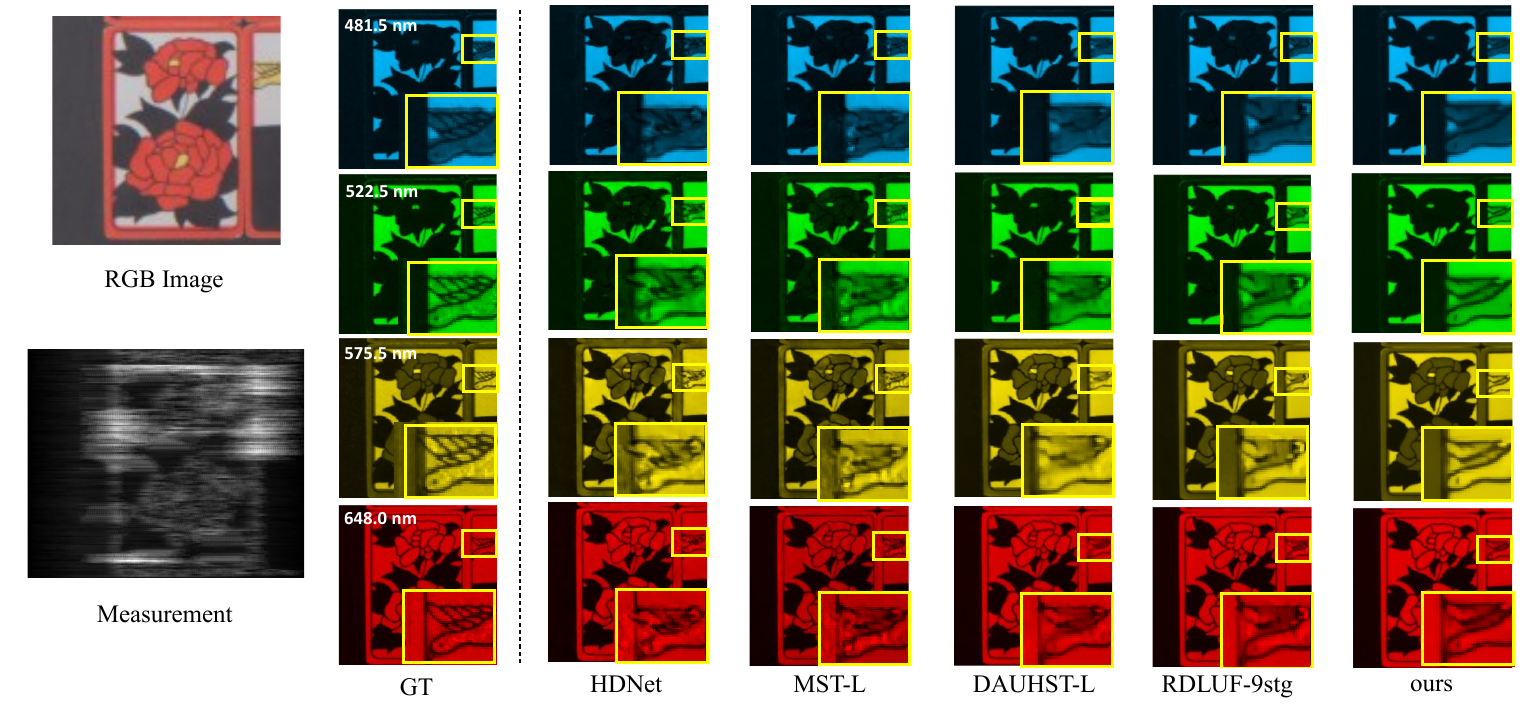}
    \caption{{Illustration of our designed adaptive step-size perception unfolding network (ASPUN).}} \label{simu}
    \end{figure*}
\textbf{NLHA.} We conducted ablation experiments on the two-branch structure and internal components of the NLHA. Specifically, the complete removal of the PNA branch results in a 0.71dB performance decrease. Removing only transformer calculation steps in PNA will also lead to performance degradation of 0.51dB. This is due to the lack of self-focused modeling of non-local pixels in transformer, and the only point convolution in GLA cannot capture non-local information, resulting in performance degradation. For the GLA branch, after completely removing this branch, ASPUN dropped 0.77dB. If only the context-aware component is removed, it is difficult for GLA to generate local perception weights, resulting in the loss of image texture and detail, and performance degradation of 0.29dB.

\textbf{Transformer Architecture in PNA.} We studied the attention architecture design used in PNA. We use ordinary window head self-attention and PNA for comparison, and their window division is shown in the figure. WMSA and PNA have similar computational costs, with PNA enjoying a larger receptive field, while WMSA perceives richer image details. We compared the two architectures with or without GLA branches, and the experimental results are shown in the table. With GLA removed, WMSA achieves slightly better performance, indicating that the receptive field advantage of PNA is limited in the design of separate branches, which is due to the absence of image detail information. In the two-branch structure with GLA, PNA achieves better performance, 0.21dB higher than WMSA. GLA brings a wealth of local detailed information, bridging the disadvantages of PNA and taking full advantage of its larger receptive field to better perceive structural information. However, in the combination of GLA and WMSA, the network's perception of local detail information is almost saturated, but it lacks attention to further pixel information, resulting in the effect of this combination is not as good as that of PNA and GLA.

\section{Conclusion}
\label{sec:prior}
In this paper, we have proposed a Adapative Step-size Perception Unfolding Network(ASPUN) based on FISTA algorithm, which uses an adaptive step perception module to estimate the update step of each channe to generate step coefficients more suitable for each spectral channel. In addition, unlike previous methods, which do not sacrifice the greatest advantage of transformer remote similarity modeling for pixel accuracy, we propose NLIA, which can take full advantage of transformers at the same computational cost. Experiment results show that ASPUN- NLHA can effectively improve hyperspectral image reconstruction performance in terms of quantity and quality, resulting in better performance over other approaches. In addition, we have recently focused on methods for sample selection in open environments. In the future, we will also further integrate deep research into open environments.



\end{document}